\documentclass{article} 
\usepackage{colm2024_conference}

\usepackage{microtype}
\usepackage{hyperref}
\usepackage{url}
\usepackage{booktabs}

\usepackage{graphicx}
\usepackage{stfloats}
\usepackage{pdflscape}

\usepackage{amsmath,amsfonts,bm}
\usepackage{wrapfig}
\usepackage{caption}
\usepackage{subcaption}

\title{Leave No Context Behind: \\ Efficient Infinite Context Transformers with Infini-attention}

\author{Tsendsuren Munkhdalai, Manaal Faruqui and Siddharth Gopal\\
Google \\
\texttt{tsendsuren@google.com}
}

\colmfinalcopy 
\begin{document}

\maketitle

\begin{abstract}
This work introduces an efficient method to scale Transformer-based Large Language Models (LLMs) to infinitely long inputs with bounded memory and computation. A key component in our proposed approach is a new attention technique dubbed Infini-attention. The Infini-attention incorporates a compressive memory into the vanilla attention mechanism and builds in both masked local attention and long-term linear attention mechanisms in a single Transformer block.
We demonstrate the effectiveness of our approach on long-context language modeling benchmarks, 1M sequence length passkey context block retrieval and 500K length book summarization tasks with 1B and 8B LLMs. Our approach introduces minimal bounded memory parameters and enables fast streaming inference for LLMs.
\end{abstract}

\section{Introduction}

Memory serves as a cornerstone of intelligence, as it enables efficient computations tailored to specific contexts. However, Transformers~\citep{vaswani2017attention} and Transformer-based LLMs~\citep{brown2020language,touvron2023llama,anil2023palm,groeneveld2024olmo} have a constrained context-dependent memory, due to the nature of the attention mechanism.
\begin{wrapfigure}{r}{0.45\textwidth}
  \begin{center}
    \includegraphics[width=0.4\textwidth]{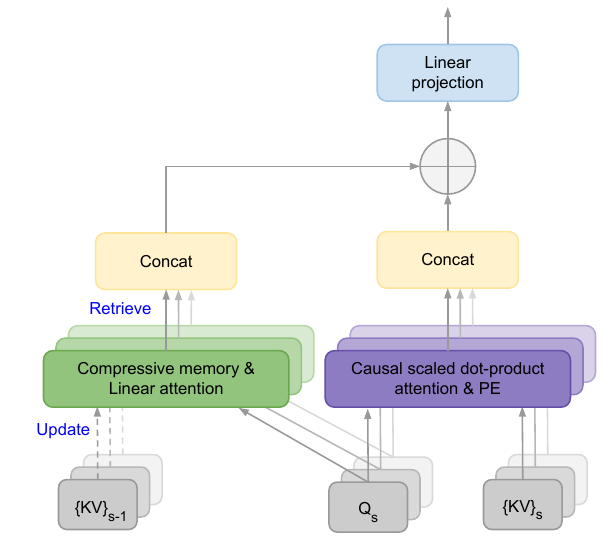}
    \caption{Infini-attention has an additional compressive memory with linear attention for processing infinitely long contexts. $\{KV\}_{s-1}$ and $\{KV\}_{s}$ are attention key and values for current and previous input segments, respectively and $Q_s$ the attention queries. PE denotes position embeddings.}
    \label{fig:infini_attention}
 \end{center}
\end{wrapfigure}
The attention mechanism in Transformers exhibits quadratic complexity in both memory footprint and computation time.
For example, the attention Key-Value (KV) states have 3TB memory footprint for a 500B model with batch size 512 and context length 2048~\citep{pope2023efficiently}.
Indeed, scaling LLMs to longer sequences (i.e. 1M tokens) is challenging with the standard Transformer architectures and serving longer and longer context models becomes costly financially.

Compressive memory systems promise to be more scalable and efficient than the attention mechanism for extremely long sequences~\citep{kanerva1988sparse,munkhdalai2019metalearned}. Instead of using an array that grows with the input sequence length, a compressive memory primarily maintains a fixed number of parameters to store and recall information with a bounded storage and computation costs. In the compressive memory, new information is added to the memory by changing its parameters with an objective that this information can be recovered back later on. However, the LLMs in their current state have yet to see an effective, practical compressive memory technique that balances simplicity along with quality.

In this work, we introduce a novel approach that enables Transformer LLMs to effectively process infinitely long inputs with bounded memory footprint and computation. A key component in our proposed approach is a new attention technique dubbed Infini-attention (Figure~\ref{fig:infini_attention}). The Infini-attention incorporates a compressive memory into the vanilla attention mechanism~\citep{bahdanau2014neural,vaswani2017attention} and builds in both masked local attention and long-term linear attention mechanisms in a single Transformer block.

Such a subtle but critical modification to the Transformer attention layer enables a natural extension of existing LLMs to infinitely long contexts via continual pre-training and fine-tuning.

Our Infini-attention reuses all the key, value and query states of the standard attention computation for long-term memory consolidation and retrieval. We store old KV states of the attention in the compressive memory, instead of discarding them like in the standard attention mechanism. We then retrieve the values from the memory by using the attention query states when processing subsequent sequences. To compute the final contextual output, the Infini-attention aggregates the long-term memory-retrieved values and the local attention contexts.

In our experiments, we show that our approach outperforms baseline models on long-context language modeling benchmarks while having ~114x comprehension ratio in terms of memory size. The model achieves even better perplexity when trained with 100K sequence length. A 1B LLM naturally scales to 1M sequence length and solves the passkey retrieval task when injected with Infini-attention. Finally, we show that a 8B model with Infini-attention reaches a new SOTA result on a 500K length book summarization task after continual pre-training and task fine-tuning.

In summary, our work makes the following contributions:
\begin{enumerate}
\item We introduce a practical and yet powerful attention mechanism – Infini-attention with long-term compressive memory and local causal attention for efficiently modeling both long and short-range contextual dependencies.
\item Infini-attention introduces minimal change to the standard scaled dot-product attention and supports plug-and-play continual pre-training and long-context adaptation by design.
\item Our approach enables Transformer LLMs to scale to infinitely long context with a bounded memory and compute resource by processing extremely long inputs in a streaming fashion.
\end{enumerate}

\section{Background}
Recurrent Neural Networks (RNNs) process a single token $x_t$ at each step $t$ and computes a recurrent hidden state $h_t$ to represent an entire input sequence~\citep{hochreiter1997long,maass2002real}:
\begin{equation}
    h_{t} = RNN(x_{t}, h_{t-1}).
\end{equation}
The RNN computation is very efficient since the model maintains only a fixed-size vector $h_t$ for input sequence. However, for processing long sequences it becomes difficult to store entire contextual information into a single fixed-size vector and this limitation had implications on RNNs utility in certain tasks~\citep{kaiser2015neural}. To address the limitation, people extended the standard RNNs with an external memory component that can be read from and written to. One such an instance is Metalearned Neural Memory (MNM)~\citep{munkhdalai2019metalearned}:
\begin{equation}
    h_{t}, \theta_{t} = MNM(x_{t}, h_{t-1}, \theta_{t-1}).
\end{equation}

MNM learns an additional memory state $\theta$ parameterized by a feed-forward neural network (FFN) and uses query, key and value vectors (QKV) to interact with the memory, similar to the attention mechanism. To store information, it modifies the parameters of the FFN by using the key vectors as input and the value vectors for the target, and to read memory entries, it forward-passes the query vectors through the memory FFN and retrieves its corresponding value. Like RNNs, the memory state is still bounded in MNM. 

Unlike the RNNs, the attention mechanism however doesn't maintain a recurrent state and only performs a feed-forward computation on input sequence segment $X_{s}$:
\begin{equation}
    O_{s} = attention(X_{s}).
\end{equation}

The attention output $O_{s}$ is simply passed to the next layer and \textbf{no state is carried over to the next input sequence ${X_{s+1}}$ at the same attention layer}. In the attention layer, in order to capture the dependency between the consequent segments $X_s$ and $X_{s+1}$, one needs to process them altogether at the same time and this process becomes a bottleneck requiring large computational resources as the length of input sequence grows more and more. To improve the efficiency while still being able to benefit from the expressiveness of the attention mechanism, \textbf{this work introduces a recurrent attention layer}.

\section{Method}

\begin{figure*}[!t]
\begin{center}
\includegraphics[width=1.0\textwidth]{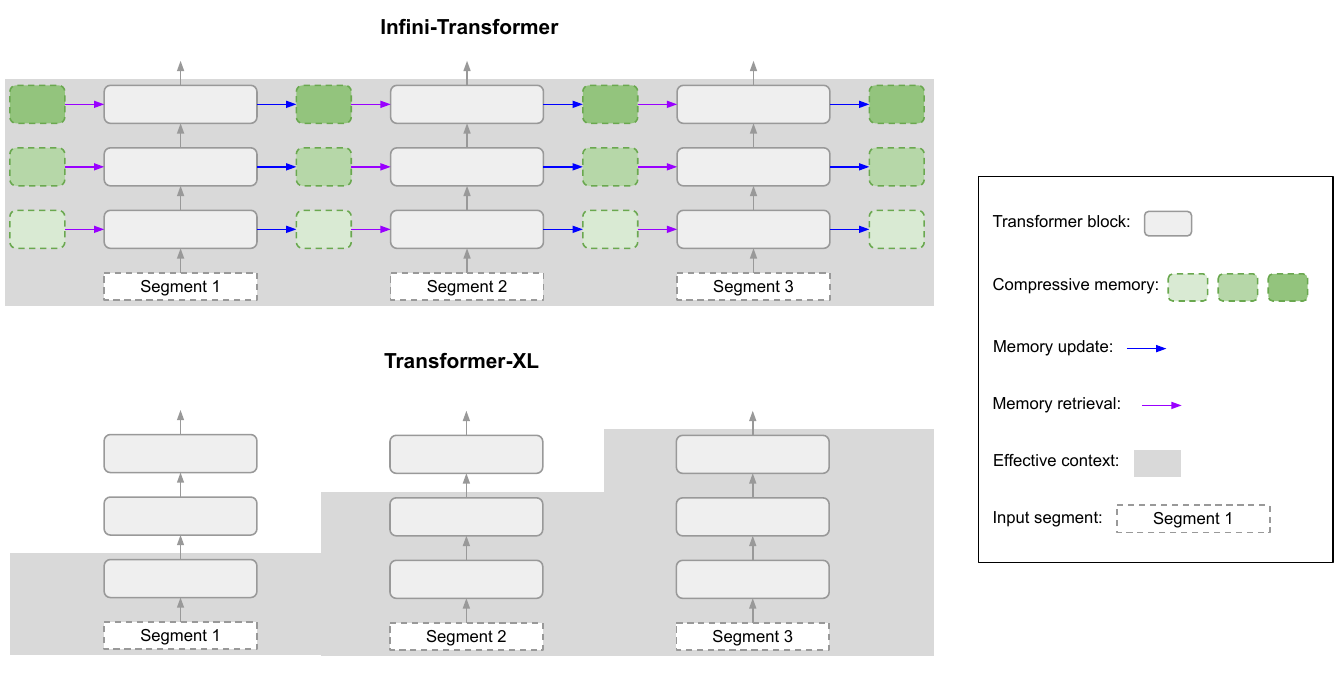}
\caption{Infini-Transformer (top) has an entire context history whereas Transformer-XL (bottom) discards old contexts since it caches the KV states for the last segment only.
}
\label{fig:infini_transformer}
\end{center}
\end{figure*}

Figure~\ref{fig:infini_transformer} compares our model,  Infini-Transformer, and Transformer-XL~\citep{dai2019transformer}.
Similar to Transformer-XL, Infini-Transformer operates on a sequence of segments. We compute the standard causal dot-product attention context within each segment. So the dot-product attention computation is local in a sense that it covers a total $N$ number of tokens of the current segment with index $S$ ($N$ is the segment length).

The local attention~\citep{dai2019transformer}, however, discards the attention states of the previous segment when processing the next one. In Infini-Transformers, instead of leaving out the old KV attention states, we propose to reuse them to maintain the entire context history with a compressive memory. So each attention layer of Infini-Transformers has both global compressive and local fine-grained states. We call such an efficient attention mechanism Infini-attention, which is illustrated in Figure~\ref{fig:infini_attention} and described formally in the following sections.

\subsection{Infini-attention}
As shown Figure~\ref{fig:infini_attention}, our Infini-attention is a recurrent attention mechanism that computes both local and global context states and combine them for its output. Similar to multi-head attention (MHA), it maintains $H$ number of parallel compressive memory per attention layer ($H$ is the number of attention heads) in addition to the dot-product attention and like the RNNs and MNM, it maintains a recurrent memory state to efficiently track the long sequence context:
\begin{equation}
    O_{s}, M_{s} = infini\text{-}attention(X_{s}, M_{s-1})
\end{equation}

\subsubsection{Scaled Dot-product Attention}
The multi-head scaled dot-product attention~\citep{vaswani2017attention}, specially its self-attention variant~\citep{munkhdalai2016citation,cheng2016long}, has been the main building block in LLMs. The MHA’s strong capability to model context-dependent dynamic computation and its conveniences of temporal masking have been leveraged extensively in the autoregressive generative models.

A single head in the vanilla MHA computes its attention context $A_{dot} \in {\rm I\!R}^{N \times d_{value}}$ from sequence of input segments $X \in {\rm I\!R}^{N \times d_{model}}$ as follows. First, it computes attention query, key, and value states:
\begin{equation}
    K = XW_K, \text{ } V = XW_V \text{ and } Q = XW_Q.
\end{equation}
Here, $W_K \in {\rm I\!R}^{d_{model} \times d_{key}}, W_V \in {\rm I\!R}^{d_{model} \times d_{value}} \text{ and } W_Q \in {\rm I\!R}^{d_{model} \times d_{key}}$ are trainable projection matrices.
Then, the attention context is calculated as a weighted average of all other values as
\begin{equation}
    A_{dot} = \text{softmax}\left(\frac{Q K^T}{\sqrt{d_{model}}}\right) V.
    \label{eq:attn}
\end{equation}

For MHA, we compute $H$ number of attention context vectors for each sequence element in parallel, concatenate them along the second dimension and then finally project the concatenated vector to the model space to obtain the attention output.

\subsubsection{Compressive Memory}
In Infini-attention, instead of computing new memory entries for compressive memory, we reuse the query, key and value states ($Q$, $K$ and $V$) from the dot-product attention computation.
The state sharing and reusing between the dot-product attention and compressive memory not only enables efficient plug-in-play long-context adaptation but also speeds up training and inference. Similar to the prior work~\citep{munkhdalai2019metalearned}, our goal is to store bindings of key and value states in the compressive memory and retrieve by using the query vectors.

While there are different forms of compressive memory proposed in the literature~\citep{hopfield1982neural,kanerva1988sparse,schlag2019enhancing,munkhdalai2019metalearned}, for simplicity and computational efficiency, in this work we parameterize the memory with an associative matrix~\citep{schlag2020learning}. This approach further allows us to cast the memory update and retrieval process as linear attention mechanism~\citep{shen2018efficient} and to leverage stable training techniques from the related methods. Specially, we adopt the update rule and retrieval mechanism by \citet{katharopoulos2020transformers} mainly due to its simplicity and competitive performance.

\textbf{Memory retrieval.} In Infini-attention, we retrieve new content $A_{mem} \in {\rm I\!R}^{N \times d_{value}}$ from the memory $M_{s-1} \in {\rm I\!R}^{d_{key} \times d_{value}}$ by using the query $Q \in {\rm I\!R}^{N \times d_{key}}$ as:
\begin{equation}
    A_{mem} =  \frac{\sigma({Q}) M_{s-1}}
    {{\sigma(Q)} z_{s-1}}.
    \label{eq:mem_ret}
\end{equation}

Here,  $\sigma$ and $z_{s-1} \in {\rm I\!R}^{d_{key}}$ are a  nonlinear activation function and a normalization term, respectively. As the choice of the non-linearity and the norm method is crucial for training stability, following \citet{katharopoulos2020transformers} we record a sum over all keys as the normalization term $z_{s-1}$ and use element-wise ELU + 1 as the activation function~\citep{clevert2015fast}.

\textbf{Memory update.} Once the retrieval is done, we update the memory and the normalization term with the new KV entries and obtain the next states as
\begin{equation}
    M_{s} \leftarrow M_{s-1} + \sigma(K)^T V \text{ and } z_{s} \leftarrow z_{s-1} + \sum_{t=1}^N \sigma(K_t).
    \label{eq:mem_update_linear}
\end{equation}
The new memory states $M_{s}$ and $z_{s}$ are then passed to the next segment $S+1$, building in a recurrence in each attention layer. The right side term $\sigma(K)^T V $ in Eq.~\eqref{eq:mem_update_linear} is known as an associative binding operator~\citep{smolensky1990tensor,hebb2005organization,schlag2020learning}.

Inspired by the success of delta rule~\citep{munkhdalai2019metalearned,schlag2020learning,schlag2021linear}, we have also incorporated it into our Infini-attention. The delta rule attempts a slightly improved memory update by first retrieving existing value entries and subtracting them from the new values before applying the associative bindings as new update.
\begin{equation}
    M_{s} \leftarrow M_{s-1} + \sigma(K)^T (V - \frac{\sigma({K}) M_{s-1}}
    {{\sigma(K)} z_{s-1}}).
    \label{eq:mem_update_delta}
\end{equation}

This update rule ($Linear + Delta$) leaves the associative matrix unmodified if the KV binding already exists in the memory while still tracking the same normalization term as the former one ($Linear$) for numerical stability.

\textbf{Long-term context injection.} We aggregate the local attention state $A_{dot}$ and memory retrieved content $A_{mem}$ via a learned gating scalar $\beta$:
\begin{equation}
    A = \textit{sigmoid} (\beta) \odot A_{mem} + (1 - \textit{sigmoid}(\beta)) \odot A_{dot}.
\end{equation}
This adds only a single scalar value as training parameter per head while allowing a learnable trade-off between the long-term and local information flows in the model~\citep{wu2022memorizing}.

Similar to the standard MHA, for the multi-head Infini-attention we compute $H$ number of context states in parallel, and concatenate and project them for the final attention output $O \in  {\rm I\!R}^{N \times d_{model}}$:
\begin{equation}
    O = [A^1; \dotso A^H] W_O
\end{equation}
where $W_O \in {\rm I\!R}^{H \times d_{value} \times d_{model}}$ is trainable weights.

\subsection{Memory and Effective Context Window}

\begin{table*}[!t]
  \centering
    \tiny
  \begin{tabular}{lllll}
    \toprule
    
    Model &	Memory (cache) footprint &	 Context length & Memory update &	Memory retrieval \\
    \hline
    Transformer-XL &	$(d_{key} + d_{value}) \times H \times N \times l$ &	$N \times l$ &	Discarded &	Dot-product attention \\
    Compressive Transformer &	$d_{model} \times (c + N) \times l$ &	$(c \times r + N) \times l$ &	Discarded &	Dot-product attention \\
    Memorizing Transformers &	$(d_{key} + d_{value}) \times H \times N \times S$ &	$N \times S$ &	None &	kNN + dot-product attention \\
    RMT &	$d_{model} \times p \times l \times 2$ &	$N \times S$ &	Discarded &	Soft-prompt input \\
    AutoCompressors &	$d_{model} \times p \times (m + 1) \times l$ &	$N \times S$ &	Discarded &	Soft-prompt input \\
    \hline
    Infini-Transformers &	$d_{key} \times (d_{value} + 1) \times H \times l$ &	$N \times S$ &	Incremental &	Linear attention \\
    \bottomrule
  \end{tabular}
  \caption{Transformer models with segment-level memory are compared. For each model, the memory size and effective context length are defined in terms of their model parameters ($N$: input segment length, $S$: the number of segments, $l$: the number of layers, $H$: the number of attention heads, $c$: Compressive Transformer memory size, $r$: compression ratio,  $p$: the number of soft-prompt summary vectors and $m$: summary vector accumulation steps).}
  \label{tab:complexity}
  \vspace{-10pt}
\end{table*}

Our Infini-Transformer enables an unbounded context window with a bounded memory footprint. To illustrate this, Table~\ref{tab:complexity} lists the previous segment-level memory models with their context-memory footprint and effective context length defined in terms of model parameters and input segment length. Infini-Transformer has a constant memory complexity of $d_{key} \times d_{value} + d_{key}$ for storing compressed context in $M_s$ and $z_{s}$ for each head in single layer while for the other models, the complexity grows along with the sequence dimension - the memory complexity depends either on the cache size for Transformer-XL~\citep{dai2019transformer}, Compressive Transformer~\citep{rae2019compressive} and Memorizing Transformers~\citep{wu2022memorizing} or on the soft-prompt size for RMT~\citep{bulatov2022recurrent} and AutoCompressors~\citep{ge2023context}.

\begin{wrapfigure}{r}{0.45\textwidth}
  \begin{center}
    \includegraphics[width=0.4\textwidth]{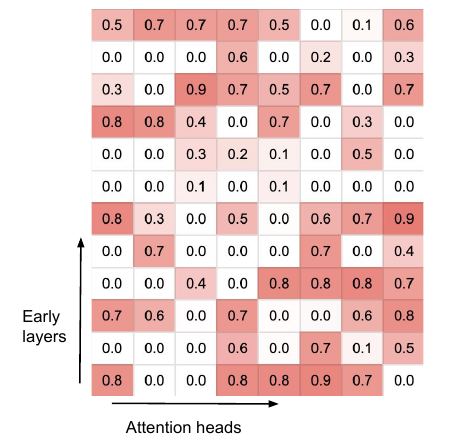}
    \caption{\small There are two types of heads emerged in Infini-attention after training: specialized heads with gating score near 0 or 1 and mixer heads with score close to 0.5. The specialized heads either process contextual information via the local attention mechanism or retrieve from the compressive memory whereas the mixer heads aggregate both current contextual information and long-term memory content together into single output.
    }
    \label{fig:memory_gate}
 \end{center}
\end{wrapfigure}

Transformer-XL computes attention over KV states cached from the last segment in addition to the current states. Since this is done for each layer, Transformer-XL extends the context window from $N$ to $N \times l$ tokens with an additional memory footprint of $(d_{key} + d_{value}) \times H \times N \times l$. Compressive Transformer adds a second cache to Transformer-XL and stores compressed representations of past segment activations. So it extends the Transformer-XL's context window by $c \times r \times l$ but still has a large context-memory complexity. Taking the idea further, Memorizing Transformers opt to store the entire KV states as context for input sequences. Since the storage becomes prohibitively expensive in this case, they restrict the contextual computation to a single layer only. By utilizing a fast kNN retriever, Memorizing Transformers then build a context window covering the entire sequence history of length $N \times S$ at an increased cost of storage. Our experiments show that Infini-Transformer LM can achieve more than 100x compression rate on top of Memorizing Transformers while further improving the perplexity score.

RMT and AutoCompressors allow for a potentially infinite context length since they compress the input into summary vectors and then pass them as extra soft-prompt inputs for the subsequent segments. However, in practice the success of those techniques highly depends on the size of soft-prompt vectors. Namely, it is necessary to increase the number of soft-prompt (summary) vectors to achieve a better performance with AutoCompressors~\citep{chevalier2023adapting} and with that, the memory and compute complexity grow quickly resulting in diminished efficiency. It was also observed in AutoCompressors~\citep{chevalier2023adapting} that an efficient compression objective is needed for training such prompt compression techniques~\citep{ge2023context}.

\section{Experiments}
We evaluated our Infini-Transformer models on benchmarks involving extremely long input sequences: long-context language modeling, 1M length passkey context block retrieval and 500K length book summarization tasks. For the language modeling benchmark, we train our models from scratch while for the passkey and book summarization tasks, we continually pre-train existing LLMs in order to highlight a plug-and-play long-context adaptation capability of our approach.

\subsection{Implementation details}
\textbf{Segment chunking.} we forward-pass the entire input text a Transformer model and then perform segment chunking at each Infini-attention layer - in this way, perform a minimal modification to the existing Transformer implementation. The Infini-attention layer segments the input and process it segment by segment and concatenates back the segments to pass the original-length segment as output to the next layer.

\textbf{Back-propagation through time (BPTT).} Each Infini-attention layer is trained with back-propagation through time~\citep{werbos1988generalization} by computing the gradient w.r.t the compressive memory states, similar to how RNNs are trained. To save memory, we perform gradient checkpoint when processing the sequence segment by segment.

\textbf{Position Embeddings (PE).} As shown Figure~\ref{fig:infini_attention}, we don't use position embeddings for the key and query vectors of the compressive memory to store only global contextual information in the long-term memory. The PEs were applied to the QK vectors only after the compressive memory reading and update.

\subsection{Long-context Language Modeling}
We trained and evaluated small Infini-Transformer models on PG19~\citep{rae2019compressive} and Arxiv-math~\citep{wu2022memorizing} benchmarks.
Our setup closely resembles that of Memorizing Transformers~\citep{wu2022memorizing}. Namely, all our models have 12 layers and 8 attention heads of dimension 128 each and FFNs with hidden layer 4096.

\begin{table*}[!t]
  \centering
    \scriptsize
  \begin{tabular}{lccccc}
    \toprule
    
    Model & Memory size (comp.) &	XL cache &	Segment length &	PG19	& Arxiv-math  \\
    \hline
    Transformer-XL & 50M (3.7x) & 2048 & 2048 & 11.88 & 2.42 \\
    Memorizing Transformers & 183M (1x) & 2048 & 2048 & 11.37 & 2.26 \\
    RMT & 2.5M (73x) & None & 2048 & 13.27 & 2.55 \\
    \hline
    Infini-Transformer\tiny{ (Linear)} & 1.6M (114x) & None & 2048 & \bf{9.65} & 2.24 \\
    Infini-Transformer\tiny{ (Linear + Delta)} & 1.6M (114x) & None & 2048 & 9.67 & \bf{2.23} \\
    \bottomrule
  \end{tabular}
  \caption{Long-context language modeling results are compared in terms of average token-level perplexity. Comp. denotes compression ratio. Infini-Transformer outperforms memorizing transformers with memory length of 65K and achieves 114x compression ratio.}
  \label{tab:lc_lm}
\end{table*}

We set the Infini-attention segment length $N$ to 2048 for all attention layers and the input sequence length to 32768 for training. This allows the Infini-attention to unroll over 16 steps w.r.t its compressive memory states. For the RMT baseline, we performed several runs with summary prompt lengths 50, 100 and 150 and sequence lengths 4096, 8196 and 32768. RMT with 100 summary vectors gave the best result when trained on 8196 length sequences.

The main results from the language modeling experiments are summarized in Table~\ref{tab:lc_lm}. Our Infini-Transformer outperforms both Transformer-XL~\citep{dai2019transformer} and Memorizing Transformers~\citep{wu2022memorizing} baselines while maintaining 114x less memory parameters than the Memorizing Transformer model with a vector retrieval-based KV memory with length of 65K at its $9^{th}$ layer.

\begin{table*}[!b]
  \centering
    \scriptsize
  \begin{tabular}{lccccc}
    \toprule
    
    {} & \multicolumn{5}{c}{ Zero-shot } \\
    \cmidrule(l{3pt}r{3pt}){2-6}
    
    & 32K & 128K & 256K & 512K & 1M \\
    \hline
    Infini-Transformer\tiny{ (Linear)}  & 14/13/98 & 11/14/100 & 6/3/100 & 6/7/99  & 8/6/98 \\
    Infini-Transformer\tiny{ (Linear + Delta)}  & 13/11/99 & 6/9/99  &   7/5/99  &  6/8/97  &  7/6/97 \\
    \toprule
    {} & \multicolumn{5}{c}{ FT (400 steps) } \\
    \cmidrule(l{3pt}r{3pt}){2-6}
    Infini-Transformer\tiny{ (Linear)} & 100/100/100 & 100/100/100 & 100/100/100 & 97/99/100 & 96/94/100 \\
Infini-Transformer\tiny{ (Linear + Delta)} & 100/100/100 & 100/100/99 & 100/100/99 & 100/100/100 & 100/100/100 \\
    \bottomrule
  \end{tabular}
  \caption{Infini-Transformers solved the passkey task with up to 1M context length when fine-tuned on 5K length inputs. We report token-level retrieval accuracy for passkeys hidden in a different part (\textit{start/middle/end}) of long inputs with lengths 32K to 1M.}
  \label{tab:passkey}
\end{table*}

\textbf{100K length training.}
We further increased the training sequence length to 100K from 32K and trained the models on Arxiv-math dataset. 100K training further decreased the perplexity score to \textbf{2.21} and \textbf{2.20} for $Linear$ and $Linear + Delta$ models.

\textbf{Gating score visualization.} Figure~\ref{fig:memory_gate} visualizes the gating score, $\textit{sigmoid}(\beta)$ for the compressive memory for all attention heads in each layer. There are two types of heads emerged in Infini-attention after training: specialized heads with a gating score near 0 or 1 and mixer heads with a score close to 0.5. The specialized heads either process contextual information via the local attention computation or retrieve from the compressive memory whereas the mixer heads aggregate both current contextual information and long-term memory content together into a single output. Interestingly, each layer has at least a single short-range head, allowing a forward-propagation of input signal up until the output layer. We also observed an interleaving of long and short-term content retrievals throughout the forward computation. 


\subsection{LLM Continual Pre-training}

\begin{table*}[!t]
  \centering
    \small
  \begin{tabular}{lcccc}
    \toprule
    
    Model & Rouge-1 &	Rouge-2 &	Rouge-L &	Overall  \\
    \hline
    BART &	36.4 &	7.6 &	15.3 &	16.2 \\
    BART + Unlimiformer &	36.8 & 8.3 &	15.7 &	16.9 \\
    PRIMERA & 38.6 &	7.2 &	15.6 &	16.3 \\
    PRIMERA + Unlimiformer &	37.9 &	8.2 &	16.3 &	17.2 \\
    \hline
    Infini-Transformers\tiny{ (Linear)} & 37.9 &	8.7 &	17.6 &	18.0 \\
    Infini-Transformers\tiny{ (Linear + Delta)} & \bf 40.0 & \bf	8.8 & \bf	17.9 &	\bf 18.5 \\
    \bottomrule
  \end{tabular}
  \caption{500K length book summarization (BookSum) results. The BART, PRIMERA and Unlimiformer results are from \citet{bertsch2024unlimiformer}.}
  \label{tab:booksum}
\end{table*}

We performed a lightweight continual pre-training for long-context adaptation of existing LLMs. The pre-training data includes the PG19 and Arxiv-math corpus as well as C4 text~\citep{raffel2020exploring} with length more than 4K tokens. The segment length $N$ was set to 2K throughout our experiments.

\textbf{1M passkey retrieval benchmark.} We replaced the vanilla MHA in a 1B LLM with Infini-attention and continued to pre-train on inputs with length of 4K. The model was trained for 30K steps with batch size of 64 before fine-tuning on the passkey retrieval task~\citep{mohtashami2024random}.

The passkey task hides a random number into a long text and asks it back at the model output. The length of the distraction text is varied by repeating a text chunk multiple times. The previous work~\citep{chen2023extending} showed that a 8B LLaMA model can solve the task up to 32K length when fine-tuned with the same 32K length inputs with Position Interpolation. We take this challenge further and fine-tune on only 5K length inputs to test on 1M length regime.

\begin{wrapfigure}{r}{0.5\textwidth}
  \begin{center}
    \includegraphics[width=0.45\textwidth]{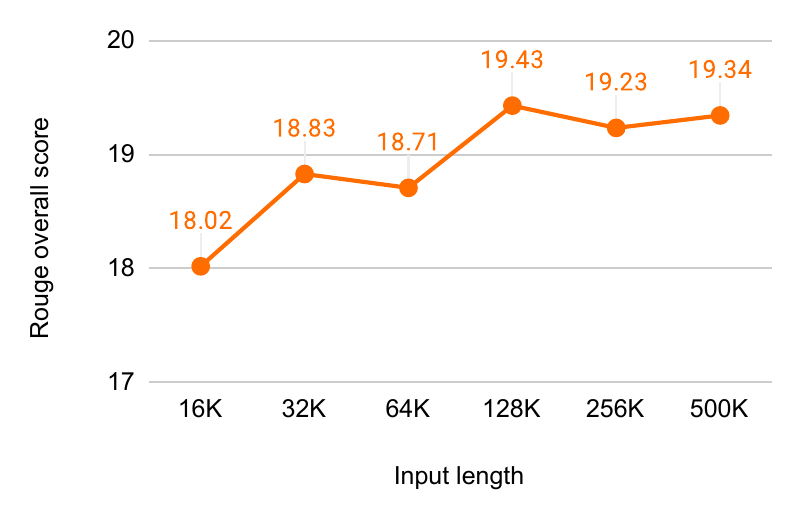}
    \caption{Infini-Transformers obtain better Rouge overall scores with more book text provided as input.}
    \label{fig:sum_lengths}
 \end{center}
\end{wrapfigure}

Table~\ref{tab:passkey} reports the token-level accuracy for test subsets with input lengths ranging from 32K to 1M. For each test subset, we controlled the position of the passkey so that it is either located around the beginning, middle or the end of the input sequence.
We reported both zero-shot accuracy and fine-tuning accuracy. Infini-Transformers solved the task with up to 1M context length after fine-tuning on 5K length inputs for 400 steps.

\textbf{500K length book summarization (BookSum).} We further scaled our approach by continuously pre-training a 8B LLM model with 8K input length for 30K steps. We then fine-tuned on a book summarization task, BookSum~\citep{kryscinski2021booksum} where the goal is to generate a summary of an entire book text.

We set the input length to 32K for fine-tuning and increase to 500K for evaluating. We use a generation temperature of 0.5 and $top_{p} = 0.95$ and set the number of decoding steps to 1024 to generate a summary of each book.

Table~\ref{tab:booksum} compares our model against the encoder-decoder models that were built particularly for the summarization task~\citep{lewis2019bart, xiao2021primera} and their retrieval-based long-context extension~\citep{bertsch2024unlimiformer}. Our model outperforms the previous best results and achieves a new SOTA on BookSum by processing the entire text from book. We have also plotted the overall Rouge score on validation split of BookSum data in Figure~\ref{fig:sum_lengths}. There is a clear trend showing that with more text provided as input from books, Our Infini-Transformers improves its summarization performance metric.

\section{Related Work}

\textbf{Compressive memory.} Inspired by the plasticity in biological neurons~\citep{munkhdalai2017meta,miconi2018differentiable}, compressive memory approaches cast parameterized functions as memory to store and retrieve information~\citep{hinton1987using,schmidhuber1992learning,ba2016using,munkhdalai2019metalearned}. Unlike the Transformer KV memory array~\citep{vaswani2017attention,wu2022memorizing}, which grows with input sequence length, compressive memory systems maintain a constant number of memory parameters for computational efficiency. The parameters are modified with an update rule to store information, which is then retrieved via a memory reading mechanism~\citep{graves2014neural,sukhbaatar2015end,munkhdalai2017neural}.

Compressed input representations can be viewed as a summary of past sequence segments~\citep{rae2019compressive,chevalier2023adapting}. Along this direction, more recent works have been utilizing a Transformer LLM itself to compress input sequence for efficient long-context modeling~\citep{bulatov2022recurrent,chevalier2023adapting,ge2023context,mu2024learning, hwang2024transformerfam}.  However, the previous segment-level compression methods, including Compressive Transformers~\citep{rae2019compressive} still discard the memory entries of old segments in order to free up space for the new ones, limiting their context window to the most recent segments. This is in contrast to our Infini-attention that computes incremental memory updates to a fixed amount of memory parameters in a recurrent fashion.

\textbf{Long-context continual pre-training.} There is a line of work that extends the dot-product attention layers and continues to train LLMs for long-context~\citep{xiong2023effective,fu2024data}. The attention extensions include incorporating sparsity into the attention layer~\citep{chen2023longlora,ratner2022parallel,mohtashami2024random} as well as manipulating the position encodings~\citep{chen2023extending,peng2023yarn}. Although the position encoding-based methods such as position interpolation techniques~\citep{chen2023extending} can be data efficient as they only adjust the positional bias in the attention layer, they are still costly for inference.

The attention mechanism is also prone to the issues of attention sink~\citep{xiao2023efficient} and lost-in-the-middle~\citep{liu2024lost}. Consequently, they struggle in a regime where context length is longer than what was observed during training~\citep{press2021train,kazemnejad2024impact}. The proposed Infini-attention addresses those issues by enabling a segment-level streaming computation over long sequences with a fixed local attention window. Our Infini-Transformers successfully extrapolate to 1M input length regimes when trained on 32K and even 5K length sequences.

\textbf{Efficient attention.} The efficient attention techniques attempt to improve the efficiency of the dot-product attention with an approximation or a system-level optimization. Multiple directions have been explored for different forms of efficient attention approximation, including sparsity-based~\citep{child2019generating,beltagy2020longformer,sukhbaatar2021not,ding2023longnet,xiao2024infllm} and linear attention approximation~\citep{shen2018efficient,katharopoulos2020transformers,schlag2021linear}.
Among those, the linear attention variants are closely related to the associative memory matrix~\citep{schlag2020learning,schlag2021linear} and the metalearned neural memory~\citep{munkhdalai2019metalearned}, where KV bindings~\citep{smolensky1990tensor} are stored in Fast-Weights~\citep{hinton1987using,schmidhuber1992learning,ba2016using} that are modified in with respect to new contextual information.
More recently, system-level optimization techniques have been proposed by leveraging specific hardware architecture to make the exact attention computation more efficient~\citep{dao2022flashattention,liu2023ring}.

\section{Conclusion}
An effective memory system is crucial not just for comprehending long contexts with LLMs, but also for reasoning, planning, continual adaptation for fresh knowledge, and even for learning how to learn. This work introduces a close integration of compressive memory module into the vanilla dot-product attention layer. This subtle but critical modification to the attention layer enables LLMs to process infinitely long contexts with bounded memory and computation resources. We show that our approach can naturally scale to a million length regime of input sequences, while outperforming the baselines on long-context language modeling benchmark and book summarization tasks. We also demonstrate a promising length generalization capability of our approach. 1B model that was fine-tuned on up to 5K sequence length passkey instances solved the 1M length problem.

\subsubsection*{Acknowledgments}
We would like to thank Dongseong Hwang for their help implementing efficient sequence unrolling mechanism with the jax scan function. We would also like to thank Aditya Gupta, Kalpesh Krishna, Tu Vu and Alexandra Chronopoulou for their feedback.

\bibliography{colm2024_conference}
\bibliographystyle{colm2024_conference}

\appendix
\section{Additional Training Details}
For the long-context language modeling task, we set the learning rate to 0.01 by performing small search over values of 0.003, 0.005, 0.01 and 0.03. We used the Adafactor optimizer~\citep{shazeer2018adafactor} with linear warmup with 1000 steps, followed by cosine decay. We applied gradient checkpointing after each segment to save to save memory. The batch size was set to 64. For the LLM experiments, we set the learning rate to 0.0001 during continual pre-training and task fine-tuning.

\section{Passkey Retrieval Task}
Below we showed the input format of the passkey task.

\sffamily
There is an important info hidden inside a lot of irrelevant text. Find it and
memorize them. I will quiz you about the important information there.
The grass is green. The sky is blue. The sun is yellow. Here we go. There and
back again. \underline{(repeat x times)} The pass key is \textbf{9054}. Remember it. \textbf{9054} is the pass key. The grass is green. The sky is blue. The sun is yellow. Here we go. There and ack again. \underline{(repeat y times)}
What is the pass key? The pass key is

\end{document}